\documentclass{article}

\usepackage[nonatbib, final]{midl_2018}

\usepackage[utf8]{inputenc} 
\usepackage[T1]{fontenc}    
\usepackage{hyperref}       
\usepackage{url}            
\usepackage{booktabs}       
\usepackage{amsfonts}       
\usepackage{amsmath}
\usepackage{nicefrac}       
\usepackage{microtype}      
\usepackage{float}
\usepackage[pdftex]{graphicx} 
\usepackage{color}
\usepackage[font=small,labelfont=bf]{caption}

\title{Training convolutional neural networks with megapixel images}

\author{
  Pinckaers, J.H.F.M. \\
  Computational Pathology Group\\
  Radboud University Medical Center\\
  Nijmegen, The Netherlands\\
  \texttt{hans.pinckaers@radboudumc.nl} \\
  \And
  Litjens, G.J.S. \\
  Computational Pathology Group\\
  Radboud University Medical Center\\
  Nijmegen, The Netherlands\\
  \texttt{geert.litjens@radboudumc.nl} \\
}

\begin{document}

\maketitle

\begin{abstract}
  To train deep convolutional neural networks, the input data and the intermediate activations need to be kept in memory to calculate the gradient descent step. Given the limited memory available in the current generation accelerator cards, this limits the maximum dimensions of the input data. We demonstrate a method to train convolutional neural networks holding only parts of the image in memory while giving equivalent results. We quantitatively compare this new way of training convolutional neural networks with conventional training. In addition, as a proof of concept, we train a convolutional neural network with 64 megapixel images, which requires 97\% less memory than the conventional approach.
\end{abstract}

\section{Introduction}
In various domains researchers work with gigapixel images which contain both global contextual information and local details, for example defense research with satellite imagery and pathology with histology slides. Training convolutional neural networks (CNNs) with those images typically involves one of two approaches: downsizing them significantly or training on image patches. The former results in a loss of local details, whereas the latter results in losing global spatial information. Additionally, when training on image patches, there needs to be labels for every patch which can be time-intensive to produce.
 
The main bottleneck of training with large images is the size of the original image that needs to be copied to the accelerator card, typically a graphics processing units (GPU), and the resulting activation maps of the network layers which need to be kept in memory for the backward pass. This puts constraints on state-of-the-art architectures, often only allowing to train with a mini-batch size of just one image. 

Recently, Gomez et al.\cite{DBLP:journals/corr/GomezRUG17} published a method to train deep residual neural networks without storing all activations, termed the Reversible Residual Network (RevNet). With RevNets only some layer activations can be recomputed from others. Another method to reduce memory usage is to recover intermediate activations by doing partial forward passes during backpropagation\cite{DBLP:journals/corr/ChenXZG16}.

We show that an intermediate activation map of a convolutional neural network can be reconstructed by doing partial forward passes with smaller parts, tiles, of the whole image up until the activation maps at a layer of choice (Figure 1b). This reconstructed activation map can then be fed as a whole through the rest of the neural network resulting in a final output. This output can then also be backpropagated to the individual tiles. We call this approach streaming stochastic gradient descent (SSGD). We will provide an implementation for SSGD and show that it is numerically equivalent to regular stochastic gradient descent. Furthermore, we will train CNN with 64 megapixel input images to show the potential of SSGD in reducing the network memory footprint.

\section{Method and Experiments}
\label{gen_inst}
To compare conventional SGD with SSGD a VGG16-like convolutional neural network was constructed as shown in Figure 1c. For the SGD experiment, this network was trained with standard settings. The same network architecture with the same parameter initialization was trained with SSGD by dividing the input image in four tiles, the activation map of the thirteenth layer was reconstructed from the result of the four tiles, after which the normal forward pass proceeded. The backpropagation can be performed from the loss until the reconstructed activation map in the traditional way. Hereafter, the gradient of the feature map is divided in relevant parts for each of the four tiles and the backpropagation is performed per tile. During backpropagation partial forward passes are used to prevent needing to keep all the tile activations in memory. We checked for numerical equivalence of the loss between the SGD and SSGD networks to ensure a correct implementation of the new forward and backward pass.

In the second experiment a CNN was trained on 64-megapixel images (8130$\times$8130). Here we divided the image in sixty-four tiles for SSGD. This experiment would require approximately 235 gigabyte GPU memory using standard SGD with a mini-batch size of 8 (29 gigabyte for mini-batch size of 1).

\textbf{Details about adapting the forward and backward passes for SSGD.} In the forward pass convolutional operations can result in smaller outputs. To preserve the same output when doing forward passes of tiles instead of full images, the outer part of the outputs that are lost or influenced by zero-padding need to be taken into account. This means that the correct amount of overlap needs to be calculated between the tiles to reconstruct the intermediate feature map. Equation 1 shows how to calculate the overlap for both the forward ($\text{P}_\text{forw}$) and backward pass ($\text{P}_\text{back}$). Let \(k\) denote the 2D kernel-size of a convolutional filter, \(b\) a Boolean which will equal 1 when we want to calculate the padding of the bottom or right-side of the output map, \(s\) the stride of a convolutional filter, and \(z\) the size of the image. Equation 1 shows how to calculate the overlap in the forward pass and this is also schematically shown in Figure 1a.

\footnotesize
\scriptsize
  \begin{equation}
  \text{P}_\text{forw} = (k - s) + b((z - k) \text{ mod } s)
  \end{equation}
  \begin{equation}
  \text{P}_\text{back} = 2(k - s) + b((z - k) \text{ mod } s)
  \end{equation}
\normalsize

For the backward pass we reconstruct the gradients of the inputs of the convolutional operations, per tile, in roughly the same way. The overlap in the backwards pass needs to be bigger since the gradient of a certain input pixel is dependent on every convolution that is applied on this pixel. For example, for a valid gradient of a pixel, for a convolution with a 3x3 kernel, two pixels on all sides of this pixel are needed to perform all the possible convolutions (see equation 2).

For layers with strides larger than one it is important that the tiles we generate are such that the same pixels are included as in the convolution of the whole image. Otherwise different results will be generated between the whole-image-based activation maps and the tile-reconstructed activation maps. To accurately deal with mini-batches we sum the gradients between images in the mini-batch and average them after the last image of a batch is processed. 

Last, in case of valid convolutions or pooling operations, it is important that the pixels lost at the right and bottom edges of tiles are taken into account to obtain equivalent results to processing the whole image. For example, this could differ if the chosen tile size is uneven and the whole image has an even size.

A full implementation of our SSGD method in PyTorch is available at: \url{https://github.com/DIAGNijmegen/StreamingSGD}.

\section{Results and discussion}
In total 564 tumor and normal images were extracted from the CAMELYON16 challenge dataset\cite{doi:10.1001/jama.2017.14585}. From the tumor mask the largest polygons were localized. Images of 8192x8192 pixels were extracted around the center point of these polygons. Images with no tumor were randomly extracted from areas with normal tissue of slides without tumor. Approximately 400 images were used as the training set, 64 as a validation set and 100 in the test set. For the first experiment in which SGD an SSGD are compared, center cropped images of 514x514 pixels were used.

\begin{figure}[h]
    \centering
    \def\svgwidth{\columnwidth}
	\includegraphics[width=\textwidth]{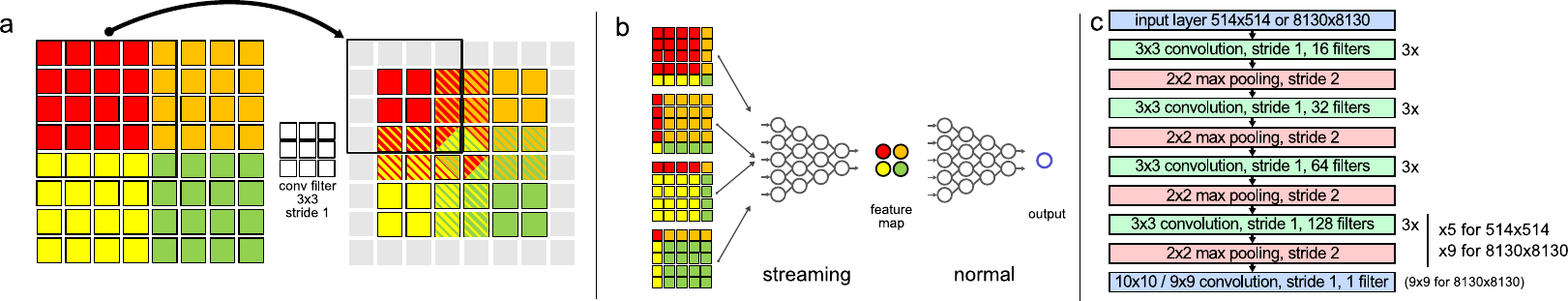}
    \caption{a) Example input and output of a 3x3 convolutional filter, arrow indicates part of the input required for certain part of the output. b) Forward pass of a streaming CNN. c) Overview of the architectures.}
\end{figure}

\begin{figure}[h]
    \centering
    \def\svgwidth{\columnwidth}
    \includegraphics[width=\textwidth]{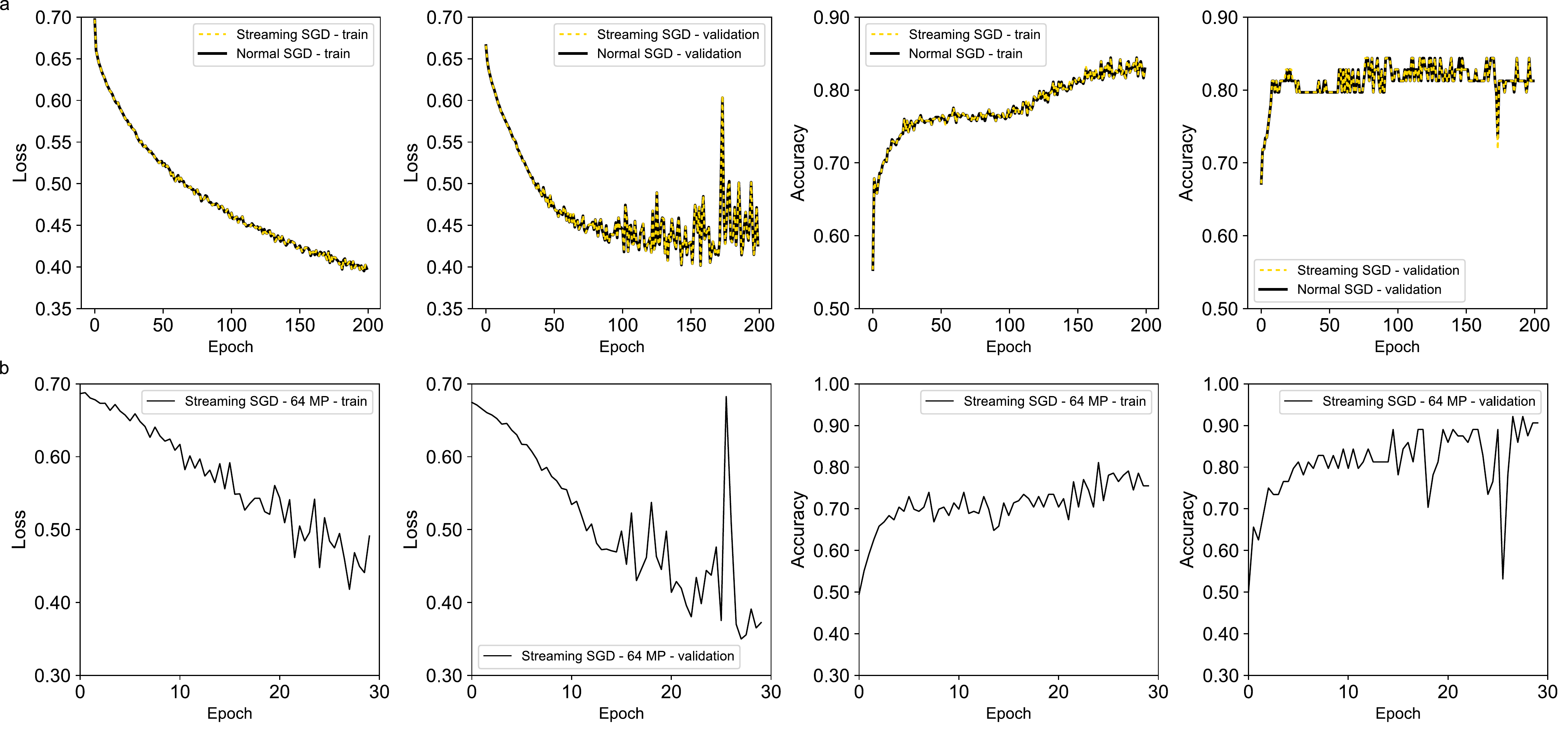}
    \caption{a) Normal SGD compared to streaming SGD. The average difference in loss was 
    $1*10^{-4}$ and in accuracy $5*10^{-4}$. Test set accuracy for normal SGD was 77\% and streaming 77\% (b) Results of the network trained on 64 megapixel (MP) images. Test accuracy was 78\%.}
\end{figure}

In the first row of Figure 2 we show the results of the quantitative comparison of the loss of training with SGD and SSGD, which is negligibly small. The small difference that is present is explained by loss of significance while calculating with small floating point numbers. This error could greatly be reduced by training with double-precision floating numbers. The second row shows that we can use SSGD to train a network with 64 megapixel images. This experiment had a memory footprint which was 97\% smaller than when training with SGD (7 GB vs. 235 GB). 

A disadvantage of SSGD is that it requires recomputation of the activation maps of the tiles during the backward pass, which can cause significant computation overhead. Beside the time trade-off, architectures are less flexible as the ability to do operations which depend on the whole activation map being present, e.g. batch normalization, cannot be used. Future directions of this method could be to further reduce memory usage in deep neural network by splitting the network on multiple levels. Another improvement would be to use RevNets in the streaming part of the network to limit the need for partial forward passes during backpropagation. Concluding, the presented SSGD method allows training of convolutional neural network with images and architectures that exceed the memory available on accelerator cards, opening up avenues for training larger networks or with larger, megapixel-sized, images.

\small
\bibliographystyle{ieeetr}
\bibliography{./references}

\begin{thebibliography}{1}

\bibitem{DBLP:journals/corr/GomezRUG17}
A.~N. Gomez, M.~Ren, R.~Urtasun, and R.~B. Grosse, ``The reversible residual
  network: Backpropagation without storing activations,'' {\em arXiv},
  vol.~abs/1707.04585, 2017.

\bibitem{DBLP:journals/corr/ChenXZG16}
T.~Chen, B.~Xu, C.~Zhang, and C.~Guestrin, ``Training deep nets with sublinear
  memory cost,'' {\em arXiv}, vol.~abs/1604.06174, 2016.

\bibitem{doi:10.1001/jama.2017.14585}
B.~E. Bejnordi, M.~Veta, and P.~J. van Diest~et al, ``Diagnostic assessment of
  deep learning algorithms for detection of lymph node metastases in women with
  breast cancer,'' {\em JAMA}, vol.~318, no.~22, pp.~2199--2210, 2017.

\end{thebibliography}

\end{document}